\documentclass{article}

\usepackage[preprint]{corl_2022} %

\usepackage[numbers]{natbib}
\usepackage{multicol}
\usepackage{amsmath}
\usepackage{amsfonts}
\usepackage{graphicx}
\usepackage{booktabs}
\usepackage{tabularx}
\usepackage{xcolor}
\usepackage{multirow}
\usepackage{gensymb}
\usepackage{xspace}
\usepackage{siunitx}
\usepackage{wrapfig}
\usepackage[bb=dsserif]{mathalpha}

\graphicspath{{figures/}}

\DeclareMathSymbol{\shortminus}{\mathbin}{AMSa}{"39}

\newcommand{\expnumber}[2]{{#1}\mathrm{e}{#2}}

\newcommand*\samethanks[1][\value{footnote}]{\footnotemark[#1]}

\newcommand{\red}[1]{}

\newcommand{\eg}{\textit{e.g.} }
\newcommand{\ie}{\textit{i.e.} }

\newcommand{\mpinet}{M\raisebox{+.1ex}{$\pi$}Nets\xspace}
\newcommand{\mpinetg}{M\raisebox{+.1ex}{$\pi$}Nets-G\xspace}
\newcommand{\mpineth}{M\raisebox{+.1ex}{$\pi$}Nets-H\xspace}
\newcommand{\mpinetc}{M\raisebox{+.1ex}{$\pi$}Nets-C\xspace}

\title{Motion Policy Networks}

\author{
Adam Fishman\thanks{Author is also affiliated with NVIDIA} \\
University of Washington \\
\texttt{afishman@cs.washington.edu} \\
\And
Adithyavairavan Murali \\
NVIDIA \\
\texttt{admurali@nvidia.com} \\
\And
Clemens Eppner \\
NVIDIA \\
\texttt{ceppner@nvidia.com} \\
\AND
Bryan Peele \\
NVIDIA \\
\texttt{bpeele@nvidia.com} \\
\And
Byron Boots\samethanks{} \\
University of Washington \\
\texttt{bboots@cs.washington.edu} \\
\And
Dieter Fox\samethanks \\
University of Washington \\
\texttt{fox@cs.washington.edu} \\
}

\begin{document}
\maketitle

\begin{abstract}
Collision-free motion generation in unknown environments is a core building block for robot manipulation. 
Generating such motions is challenging due to multiple objectives; not only should the solutions be optimal, the motion generator itself must be fast enough for real-time performance and reliable enough for practical deployment.
A wide variety of methods have been proposed ranging from local controllers to global planners, often being combined to offset their shortcomings. We present an end-to-end neural model called Motion Policy Networks (\mpinet) to generate collision-free, smooth motion from just a single depth camera observation. \mpinet~are trained on over \num{3}~million motion planning problems in \num{500000} environments. 
Our experiments show that \mpinet~ are significantly faster than global planners while exhibiting the reactivity needed to deal with dynamic scenes. They are $46\%$ better than prior neural planners and more robust than local control policies. Despite being only trained in simulation, \mpinet~transfer well to the real robot with noisy partial point clouds. Videos and code are available at \url{https://mpinets.github.io}.

\end{abstract}


\begin{figure*}[!ht]
  \centering
  \includegraphics[width=\textwidth]{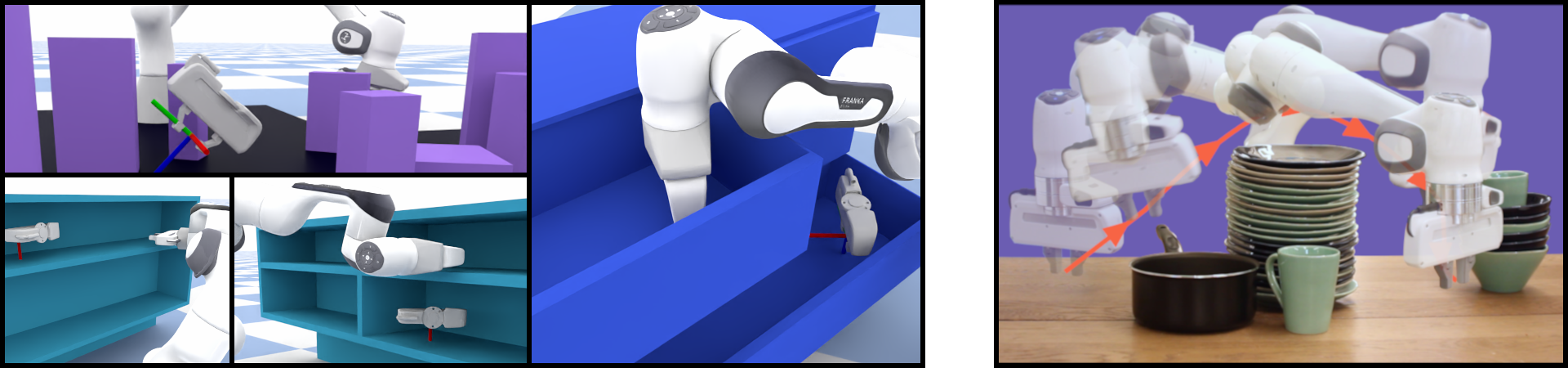}
  \caption{\mpinet~are trained on a large dataset of synthetic demonstrations~(\textit{left}) and can solve complex motion planning problems using raw point cloud observations~(\textit{right}).}
  \label{fig:dataset}
  
\end{figure*}
\section{Introduction}
Generating fast and legible motions for a robotic manipulator in unknown environments is still an open problem. Decades of research have established many well-studied algorithms, but there are two practical issues that prevent motion planning methods from being widely adopted in industrial applications and home environments that require real-time control. First, it is challenging for any single approach to satisfy multiple planning considerations: speed, completeness, optimality, ease-of-use, legibility (from the perspective of a human operator), determinism, and smoothness. Second, existing approaches enforce strong assumptions about the visual obstacle representations---such as accurate collision checking in configuration space \cite{LaValle2006PlanningA} or the availability of a gradient \cite{Ratliff2009CHOMPGO, Schulman2014MotionPW, VanWyk2022GeometricFG}---and hence require intermediate processing to operate in novel scenes directly using raw sensor observations.

Global planners such as RRT~\cite{LaValle1998RapidlyexploringRT} are useful to quickly find a feasible path but say nothing about optimality. Other sampling-based approaches iteratively refine their paths to reduce cost and asymptotically approach the optimal solution~\cite{Karaman2011SamplingbasedAF,Gammell2015BatchIT,Strub2020AdvancedB,Strub2020AdaptivelyIT}.
Optimization-based approaches~\cite{Ratliff2009CHOMPGO,Schulman2014MotionPW,Mukadam-IJRR-18} embrace locally optimal behavior in exchange for completeness. Recent methods such as Geometric Fabrics~\cite{VanWyk2022GeometricFG} and STORM~\cite{Bhardwaj2021FastJS} deploy reactive local policies and assume that local decisions will lead to globally acceptable paths. Unfortunately, as we show in our experiments, the performance of these local approaches degrades in more geometrically complex environments as they get stuck in local minima. Motivated by the success of deep learning, neural motion planning approaches such as Motion Planning Networks~\cite{Qureshi2019MotionPN} have been proposed to greatly improve the sampling of an RRT planner with imitation learning. However, they still require a planner and a collision checker with known models at test time.

Planners have traditionally been evaluated with known environment models and perfect state estimation. When deploying them in practice, however, one would have to create one of several scene representations: a static or dynamic mesh, occupancy grids~\cite{Ichter2018LearningSD,kew2019neural}, signed distance fields, etc. Reconstruction systems such as SLAM and KinectFusion~\cite{KinectFusion2011} have a large system start-up time, require a moving camera to aggregate many viewpoints, and ultimately require costly updates in the presence of dynamic objects. Recent implicit deep learning methods like DeepSDF~\cite{DeepSDF2019JJPark} and NeRF~\cite{Mildenhall2020NeRFRS} are slow or do not yet generalize to novel scenes. Methods such as SceneCollisionNet~\cite{Danielczuk2021ObjectRU} provide fast collision checks but require expensive MPC rollouts at test time. It also draws samples from a straight line path in configuration space which may not generalize to challenging environments beyond a tabletop. Other RL-based methods learn a latent representation from observations but have only been applied to simple 2D~\cite{JurgensonRSS2019, Tamar2016VIN} or 3D~\cite{Strudel2020LearningOR} environments in simulation. 

We present \textit{Motion Policy Networks (\mpinet)}, a novel method for learning an end-to-end policy for motion planning. Our approach circumvents the challenges of traditional motion planning and is flexible enough to be applied in unknown environments.
Our contributions are as follows:

\vspace{-1mm}
\begin{list}{\textbullet}{\leftmargin=2em \itemindent=0em}
    \item We present a large-scale effort in neural motion planning for manipulation. Specifically, we learn from over \num{3} million motion planning problems across over \num{500000} instances of three types of environments, nearly $300$x larger than prior work \cite{Qureshi2019MotionPN}.
    \item We train a reactive, end-to-end neural policy that operates on point clouds of the environment and moves to task space targets while avoiding obstacles. Our policy is significantly faster than other baseline configuration space planners and succeeds more than local task space controllers.
    \item On our challenging dataset benchmarks, we show that \mpinet is nearly $46\%$ more successful at finding collision-free paths than prior work \cite{Qureshi2019MotionPN} without even needing the full scene collision model.
    \item Finally, we demonstrate \textit{sim2real} transfer to real robot partial point cloud observations.
\end{list}
\section{Related Work}
\label{sec:related-work}
\textbf{Global Planning:} Robotic motion planning typically splits into three camps: search, sampling, and optimization-based planning. Search-based planning algorithms, such as A*~\cite{Hart1968AFB, Likhachev2003ARAAA, Likhachev2005AnytimeDA}, discretize the state space and perform a graph search to find an optimal path. While the graph search can be fast, complete, and guaranteed optimal, the requirement to construct a discrete graph hinders these algorithms in continuous spaces and for novel problems not well covered by the current graph. Sampling-based planners~\cite{LaValle1998RapidlyexploringRT} function in a continuous state space by drawing samples and building a tree. When the tree has sufficient coverage of the planning problem, the algorithm traverses the tree to produce the final plan. Sampling based planners are continuous, probabilistically complete, \ie find a solution with probability \num{1}, and some are even \textit{asymptotically optimal}~\cite{Karaman2011SamplingbasedAF, Gammell2015BatchIT, Strub2020AdvancedB}, but under practical time limitations,  their random nature can produce erratic\textemdash though valid\textemdash paths.

Both of the aforementioned planner types are designed to optimize for path length in the given state space (\eg configuration space) while avoiding collisions. An optimal path in configuration space is not necessarily optimal for the end effector in cartesian space. Humans motion tends to minimize hand distance traveled \cite{Uno1989FormationAC}, so what appears optimal for the algorithm may be unintuitive for a human partner or operator.
In the manipulation domain, goals are typically represented in end effector task space \cite{Qureshi2021NeRPNR, Sundermeyer2021ContactGraspNetE6}. In a closed loop setting with a moving target, the traditional process of using IK to map task to configuration space can produce highly variable configurations, especially around obstacles.
Motion Optimization \cite{Ratliff2009CHOMPGO, Schulman2014MotionPW, Ratliff2015UnderstandingTG} on the other hand, generates paths with non-linear optimization and can consider multiple objectives such as smoothness of the motion, obstacle avoidance and convergence to an end effector pose. These algorithms require careful tuning of the respective cost functions to ensure convergence to a desirable path and are prone to local minima. Furthermore, non-linear optimization is computationally complex and can be slow for difficult planning problems.

\textbf{Local Control:} In contrast to global planners, local controllers have long been applied to create collision-free motions~\cite{khatib1986real, Ratliff2018RiemannianMP, VanWyk2022GeometricFG, Bhardwaj2021FastJS}. While they prioritize speed and smoothness, they are highly local and may fail to find a valid path in complex environments. We demonstrate in our experiments that \mpinet are more effective at producing convergent motions in these types of environments, including in dynamic and in partially observed settings.

\textbf{Imitation Learning:}
Imitation Learning \cite{Osa2018AnAP} can train a policy from expert demonstrations with limited knowledge of the expert's internal model. For motion planning problems, we can apply imitation learning and leverage a traditional planner as the expert demonstrator---with perfect model of the scene during training---and learn a policy that forgoes the need for an explicit scene model at test time. Popular imitation learning methods include Inverse Reinforcement Learning \cite{Russell1998LearningAF, Ng2000AlgorithmsIRL, Ziebart2008MaximumEI} and Behavior Cloning \cite{Pomerleau1988ALVINNAA, Bain1995AFF}. The former typically assumes expert optimality and learns a cost function accordingly, whereas the latter directly learns the state-action mapping from demonstrations, regardless of the expert's optimality. We thus employ behavior cloning because producing optimal plans for continuous manipulation problems is challenging. Recent work demonstrates behavior cloning's efficacy for fine-grained manipulation tasks, such as chopstick use \cite{Ke2021GraspingWC} and pick-and-place \cite{robomimic2021}. For long-horizon tasks like ours, however, distributional shift and data variance can hinder behavior cloning performance. Distribution shift during execution can lead to states unseen in training data \cite{Ke2021GraspingWC}. Complex tasks often have a long tail of possible action states that are underrepresented in the data, leading to high data variance \cite{Codevilla2019ExploringTL}. There are many techniques to address these challenges through randomization, noise injection, regret optimization, and expert correction \citep{Ke2021GraspingWC, Ross2011ARO, ross2010A, Laskey2017DARTNI, ross2010B}. These techniques, however, have not been demonstrated on a problem of our scale and complexity (see Appendix \ref{app:data-generation} for details on the range of data). Our proposal seeks to overcome these issues by specifically designing a learnable expert, increasing the scale and variation of the data, and using a sufficiently expressive policy model.

\textbf{Neural Motion Planning:} Many deep planning methods \cite{Ichter2018LearningSD, Kumar2019LEGOLE, Zhang2018LearningIS, Chamzas2021LearningSD} seek to learn efficient samplers to speed up traditional planners. Motion Planning Networks (MPNets)~\cite{Qureshi2019MotionPN} learn to directly plan through imitation of a standard sampling based RRT* planner~~\citep{Karaman2011SamplingbasedAF} and is used in conjunction with a traditional planner for stronger guarantees. While these works greatly improve the speed of the planning search, they have the same requirements as a standard planning system: targets in configuration space and an explicit collision checker to connect the path. Our work operates based on task-space targets and perceptual observations from a depth sensor without explicit state estimation.

Novel architectures have been proposed, such as differentiable planning modules in Value Iteration Networks \cite{Tamar2016VIN}, transformers by \citet{Singh2021SPT} and goal-conditioned RL policies \cite{Eysenbach2021CLearning}. These methods are challenging to generalize to unknown environments or have only been shown in simple 2D \cite{JurgensonRSS2019} or 3D settings \cite{Strudel2020LearningOR}. In contrast, we illustrate our approach in the challenging domain of controlling a $7$ degrees of freedom~(DOF) manipulator in unknown, dynamic environments.

\section{Learning from Motion Planning}

\subsection{Problem Formulation}
\mpinet~expect two inputs, a robot configuration $q_t$ and a segmented, calibrated point cloud $z_t$. Before passing $q_t$ through the network, we normalize each element to be within $[-1, 1]$ according to the limits for the corresponding joint. We call this $q_t^{\lVert\cdot\rVert}$. The point cloud is always assumed to be calibrated in the robot's base frame, and it encodes three segmentation classes: the robot's current geometry, the scene geometry, and the target pose. 
Targets are inserted into the point cloud via points sampled from the mesh of a floating end effector placed at the target pose.  

The network produces a displacement within normalized configuration space $\dot{q}_t^{\lVert\cdot\rVert}$. To get the next predicted state $\hat{q}_{t+1}$, we take $q_t^{\lVert\cdot\rVert} + \dot{q}_t^{\lVert\cdot\rVert}$, clamp between $[-1, 1]$, and unnormalize. During training, we use $\hat{q}_{t+1}$ to compute the loss, and when executing, we use $\hat{q}_{t+1}$ as the next position target for the robot's low-level controller. 

\subsection{Model Architecture}
\label{sec:learning-architecture}
\begin{figure}[!ht]
 \centering
 \includegraphics[width=\columnwidth]{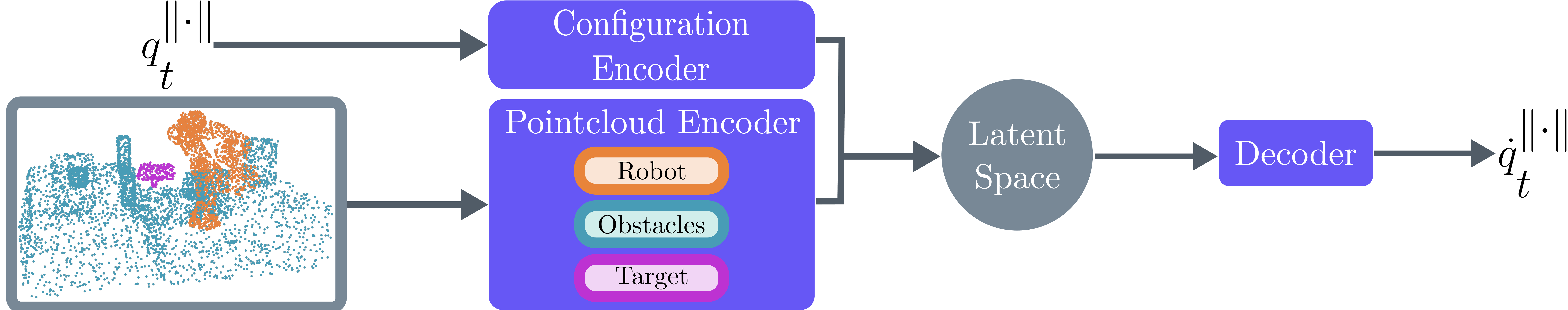}
 \caption{\mpinet~encodes state as a normalized robot configuration and segmented point cloud with three classes for the robot, the obstacles, and the target. The policy outputs a displacement in normalized joint space, which can then be applied to the input before unnormalizing to get $q_{t+1}$.}
 \label{fig:architecture}
\end{figure}

The network consists of two separate encoders, one for the point cloud and one for the robot's current configuration, as well as a decoder, totaling $19$M parameters. Our neural policy architecture is visualized in Fig. \ref{fig:architecture}.
We use PointNet++ \cite{Qi2017PointNetDH} for our point cloud encoder. PointNet++ learns a hierarchical point cloud representation and can encode a point cloud's 3D geometry, even with high variation in sampling density. PointNet++ architectures have been shown to be effective for a variety of point cloud processing tasks, such as segmentation \cite{Qi2017PointNetDH}, collision checking \cite{Danielczuk2021ObjectRU}, and robotic grasping \cite{Mousavian2018GraspNet, Murali2020CollisionNet}. Additionally, PointNet++ includes PointNet as a subcomponent. PointNet is effective at processing partially observed pointclouds, even when trained exclusively with fully-observed scenes \cite{Qi2017PointNetDL}. The robot configuration encoder and the displacement decoder are both fully connected multilayer perceptrons. We discuss the architecture in detail in Appendix \ref{app:network-architecture}.

\subsection{Loss Function}
The network is trained with a compound loss function with two constituent parts: a behavior cloning loss to enforce accurate predictions and a collision loss to safeguard against catastrophic behavior.

\paragraph{Geometric Loss for Behavior Cloning}\label{eqn:alignment_loss} To encourage alignment between the prediction and the expert, we compute a geometric loss across a set of \num{1024} fixed points along the surface of the robot.
\begin{equation}
L_\text{BC}(\hat{\Delta} q_t) = \sum_i \lVert \hat{x}^i_{t+1} - x^i_{t+1}\rVert_2 + \lVert \hat{x}^i_{t+1} - x^i_{t+1}\rVert_1 \text{, where}\;\; \begin{aligned}
\hat{x}^i_{t+1} &= \phi^i(q_t + \hat{\Delta}q_t) \\
x^i_{t+1} &= \phi^i(q_{t+1})
\end{aligned}
\end{equation}

$\phi^i(\cdot)$ represents a forward kinematics mapping from the joint angles of the robot to point~$i$ defined on the robot's surface. 
The loss is computed as the sum of the $L1$ and $L2$ distances between corresponding points on the expert and the prediction after applying the predicted displacement. By using both $L1$ and $L2$, we are able to penalize both large and small deviations.

We use a geometric, task-space loss because our goal is to ensure task-space consistency of our policy. Configuration space loss appears in prior work~\cite{Qureshi2019MotionPN}, but does not capture the accumulated error of the kinematic chain as effectively (see Appendix \ref{app:additional-experiments}).

\paragraph{Collision Loss}\label{eqn:sdf_loss} In order to avoid collisions--a catastrophic failure--we apply an additional hinge-loss inspired by motion optimization \cite{Fishman2020CollaborativeIM}. 
\begin{equation}
    L_\text{collision} = \sum_i \sum_j \lVert h_j(\hat{x}^i_{t+1}) \rVert_2\text{, where}\; h_j(\hat{x}^i_{t+1}) = \begin{cases}
-D_j(\hat{x}^i_{t+1}), & \text{if } D_j(\hat{x}^i_{t+1}) \leq 0 \\
0, & \text{if } D_j(\hat{x}^i_{t+1}) > 0 
\end{cases} 
\end{equation}

The synthetic environments are fully-observable during training, giving us access to the signed-distance functions (SDF), $\left\{D_j(\cdot)\right\}_j$, of the obstacles in each scene. For a given closed surface, its SDF maps a point in Euclidean space to the minimum distance from the point to the surface. If the point is inside the surface, the function returns a negative value.

\subsection{Training Implementation Details}
\label{subsec:data-augmentation}

\mpinet is trained for single-step prediction, but during inference, we use it recursively for closed-loop rollouts. The compounded noise in subsequent inputs equates covariate shift~ \cite{ross2010A, ross2010B}. To promote robustness, we augment our training data with random perturbations in two ways. We apply Gaussian noise to the joint angles of each input configuration, which in turn affects the corresponding points in the point cloud, passed as input to the network~\cite{Ke2021GraspingWC, DART2017Laskey}. Additionally, for each training example, we generate a unique point cloud during training, \ie during each epoch, the network sees \num{163.5}M unique point clouds. We train our network with a single set of weights across our entire dataset. 

\section{Procedural Data Generation}
\begin{figure}[!ht]
 \centering
 \includegraphics[width=\columnwidth]{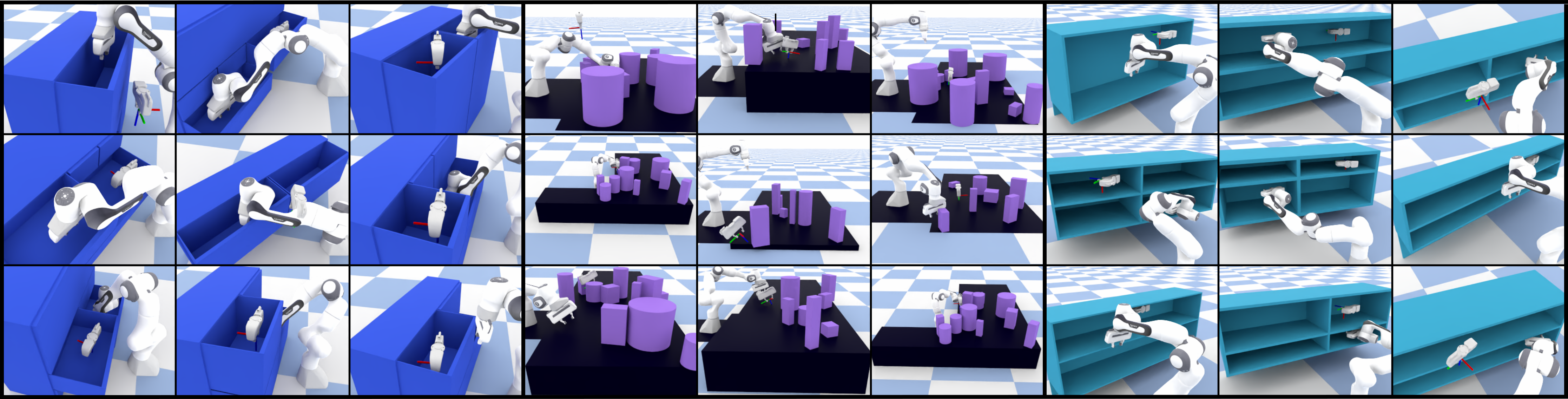}
 \caption{\mpinet is trained with a dataset consisting of solutions to $3.27$~million unique planning problems across over \num{575000}~unique, procedurally generated environments.}
 \label{fig:environments}
\end{figure}

\subsection{Large-scale Motion Planning Problems}
\label{subsec:data-problems}
Each planning problem is defined by three components: the scene geometry, the start configuration, and the goal pose. Our dataset consists of randomly generated problems across all three components, totaling $3.27$~million problems in over $575,000$~environments.
We have three classes of problems of increasing difficulty: a cluttered tabletop with randomly placed objects, cubbies and dressers. Representative examples of these environments are shown in Fig. \ref{fig:dataset}. Once we build these environments, we generate a set of potential end-effector targets and corresponding inverse kinematics solutions. We then randomly choose pairs of these configurations and verify if a plan exists between them using our expert pipeline, as detailed further in Sec. \ref{subsec:data-plans} and in the Appendix \ref{app:data-generation}.

\subsection{Expert Pipeline}
\label{subsec:data-plans}
Our expert pipeline is designed to produce high quality demonstrations we want to mimic, \ie trajectories with smooth, consistent motion and short path lengths. Here, \textit{consistency} is meant to describe quality and repeatability of an expert planner---see Appendix \ref{app:expert-pipelines} for further discussion. We considered two candidates for the expert - the \textit{Global Planner} which is a typical state-of-the-art configuration space planning pipeline \cite{Strub2020AdaptivelyIT} and a \textit{Hybrid Planner} that we engineered specifically to generate consistent motion in task space. For both planners, we reject any trajectories that produce collisions, exceed the joint limits, exhibit erratic behavior (\ie high jerk), or that have divergent motion (\ie final task space pose is more than $5\si{\cm}$ from the target). 

\textbf{Global Planner} consists of off-the-shelf components of a standard motion planning pipeline--inverse kinematics (IK)~\cite{diankov_thesis}, configuration-space AIT*~\cite{Strub2020AdaptivelyIT}, and spline-based, collision-aware trajectory smoothing~\cite{Hauser2010FastSO}. 
For a solvable problem, as the planning time approaches infinity, IK will find a valid solution and AIT* will produce an optimal collision-free path, both with probability \num{1}. Likewise, with continuous collision checking, the smoother will produce a smooth, collision-free path. In practice, 
our dataset size goal---we generated \num{6.54}M trajectories across over \num{773}K environments---dictated our computation budget and we tuned the algorithms according to this limit. 
We attempted IK at most \num{1000} times, utilized an AIT* time out of \num{20}\si{\second}, and employed discrete collision checking when smoothing. Most commonly, the pipeline failed when AIT* timed out or when, close to obstacles, the smoother's discrete checker missed a collision, thereby creating invalid trajectories.

\textbf{Hybrid Planner} is designed to produce consistent motion in task space. The planner consists of task-space AIT* \cite{Strub2020AdaptivelyIT} and Geometric Fabrics \cite{VanWyk2022GeometricFG}. AIT* produces an efficient end effector path and Geometric Fabrics produce geometrically consistent motion. The end effector paths acts as a dense sequence of waypoints for a sequence of Geometric Fabrics, but as the robot moves through the waypoints, the speed can vary. To promote smooth configuration space velocity over the final trajectory, we fit a spline to the path and retime it to have steady velocity. As we discuss in Sec. \ref{sec:comparison-state}, Geometric Fabrics often fail to converge to a target, so we redefine the planning problem to have the same target as the final position of the trajectory produced by the expert. Inspired by \cite{Andrychowicz2017HindsightER}, we call this technique \textit{Hindsight Goal Revision (HGR)} and demonstrate its importance in Sec. \ref{subsec:ablations}. Using the \textit{Hybrid Planner}, we generated $3.27$ million trajectories across \num{576532} environments.

\section{Experimental Evaluation}
\label{sec:results}
We evaluate our method with problems generated from the same distribution as the training set. See Appendix~\ref{app:data-generation} for more detail on the procedural generation and random distribution. Within the test set, each problem has a unique, randomly generated environment, as well as a unique target and starting configuration. None of the test environments, starting configurations, nor goals were seen by the network during training. Our evaluations were performed on three test sets: a set of problems solvable by the \textit{Global Planner}, problems solvable by the \textit{Hybrid Planner}, and problems solvable by both. Each test set has \num{1800} problems, with \num{600} in each of the three types of environments. 

\textbf{Quantitative Metrics:} To understand the performance of a policy, we roll it out until it matches one of two termination conditions: 1) the Euclidean distance to the target is within 1cm or 2) the trajectory has been executed for \SI{20}{\second} (based on consultations with the authors of \cite{VanWyk2022GeometricFG} and \cite{Bhardwaj2021FastJS}). We consider the following metrics (see Appendix~\ref{app:metrics} for details):
\begin{list}{\textbullet}{\leftmargin=2em \itemindent=0em}
    \item \textit{Success Rate} - A trajectory is considered a success if its final position and orientation target errors are below~\SI{1}{\cm} and \SI{15}{\degree} respectively and there are no physical violations.
    \item \textit{Time} - We measure the wall time for each \textit{successful} trajectory. We also measure \textit{Cold Start (CS) Time}, the average time to react to a new planning problem.
    \item \textit{Rollout Target Error} - The L$2$ position and orientation error (taken from \cite{Wunsch1997RealtimePE}) between the target and final end-effector pose in the trajectory.
    \item \textit{Collision Rate} - The rate of fatal collisions, both self and scene collisions
    \item \textit{Smoothness} - We use Spectral Arc Length~(SPARC)~\cite{Balasubramanian2015OnTA} and consider a path to be smooth if its SPARC values in joint and end-effector space are below \SI{-1.6}{}.
\end{list}

\begin{table*}\centering\footnotesize
\addtolength{\tabcolsep}{-2pt}    
\begin{tabular}{@{}lcccccc@{}}
\toprule
& & & \multicolumn{3}{c}{Success Rate ($\%$)} \\
\cmidrule(lr){4-6}
    & Soln. Time (\si{\second}) & CS Time (\si{\second}) & Global & Hybrid & Both & Smooth ($\%$) \\
\midrule  
\addlinespace
Global Planner \cite{Strub2020AdaptivelyIT} & $16.46 \pm 0.90$ & $16.46 \pm 0.90$ & $100$ & $78.44$ & $100$ & $51.00$ \\
Hybrid Planner & $7.37 \pm 2.23$ & $7.37 \pm 2.23$ & $50.22$ & $100$ & $100$ & $99.26$ \\
\addlinespace
G. Fabrics \cite{VanWyk2022GeometricFG} & $0.15 \pm 0.09$ & $\expnumber{2.4}{\shortminus4} \pm \expnumber{3}{\shortminus5}$ & $38.44$ & $59.33$ & $60.06$ & $85.39$ \\ 
STORM \cite{Bhardwaj2021FastJS} & $4.03 \pm 1.89$ & $\expnumber{13.4}{\shortminus3} \pm \expnumber{2.2}{\shortminus3}$ & $50.22$ & $74.50$ & $76.00$ & $62.26$ \\ 
\addlinespace
MPNets~\cite{Qureshi2019MotionPN} \\
\;\;\textit{Hybrid Expert} & $4.95 \pm 23.51$ & $4.95 \pm 23.51$ & $41.33$ & $65.28$ & $67.67$ & $99.97$ \\
\;\;\textit{Random} & $0.31 \pm 3.55$ & $0.31 \pm 3.55$ & $32.89$ & $55.33$ & $58.17$ & $99.96$ \\
\addlinespace
\mpinet (Ours) \\  
    \;\;\textit{Global Expert} & $0.33 \pm 0.08$ & $\expnumber{6.8}{\shortminus3} \pm \expnumber{7}{\shortminus5}$ & $75.06$ & $80.39$ & $82.78$ & $89.67$ \\ 
    \;\;\textit{Hybrid Expert} & $0.33 \pm 0.08$ & $\expnumber{6.8}{\shortminus3} \pm \expnumber{7}{\shortminus5}$ & $75.78$ & $95.33$ & $95.06$ & $93.81$ \\ 
\bottomrule  
\end{tabular}  
\caption{Algorithm performance on problems sets solvable by planner types. All prior methods use state-information and a oracle collision checker while \mpinet only needs a point cloud}  
\label{tab:success-rates}
\addtolength{\tabcolsep}{2pt}
\end{table*}  

\subsection{Comparison to Methods With Complete State}
\label{sec:comparison-state}
Most methods to generate motion in the literature assume access to complete state information in order to perform collision checks. In each of the following experiments, we provide each baseline method with an oracle collision checker. When running \mpinet, we use a point cloud sampled uniformly from the surface of the entire scene. Results are shown in Table \ref{tab:success-rates}. 

\paragraph{Global Configuration Space Planner}
The \textit{Global Planner} is unmatched in its ability to reach a target, but this comes at the cost of average computation time ($16.46\si{\second}$) compared to \mpinet ($0.33\si{\second}$). With a global planner, there is no option to partially solve a problem, meaning the Cold Start Time is equal to the planning time. In a real system, optimizers~\citep{Ratliff2009CHOMPGO, Schulman2014MotionPW, Mukadam-IJRR-18} could be used to quickly replan once an initial plan has been discovered. 
As discussed in Sec. \ref{subsec:data-plans}, the \textit{Global Planner} is theoretically complete, but fails in practice on some of the \textit{Hybrid Planner}-solvable problems due to system timeouts and discrete collision checking during smoothing. 

\paragraph{Hybrid End-Effector Space Planner}
Our \textit{Hybrid Planner} struggles with a large proportion of problems solvable by the \textit{Global Planner}. Yet, its solutions are both faster and smoother than the \textit{Global Planner}. Surprisingly, \mpinet trained with data from the expert \emph{outperformed} the expert on the \textit{Global Planner}-solvable test set. We attribute this to two features: 1) we use strict rejection sampling to reduce erratic and divergent behavior in our expert dataset and train only on the filtered data and 2) our use of Hindsight Goal Revision to turn an imperfect expert into a perfect one.

\begin{table*}\centering\footnotesize
\begin{tabular}{@{}lccc@{}}
\toprule
& & \multicolumn{2}{c}{Evaluation Set} \\
\cmidrule(lr){3-4} 
    & Training Set & MPNets-Style & Hybrid Expert Solvable (Ours) \\
\midrule  
\addlinespace
MPNets~\cite{Qureshi2019MotionPN} & MPNets-Style & $78.70$ & $49.89$ \\
\mpinet (Ours) & MPNets-Style & $33.70$ & $5.50$ \\
\addlinespace
MPNets~\cite{Qureshi2019MotionPN} & Hybrid Expert & $88.90$ & $65.28$ \\
\mpinet (Ours) & Hybrid Expert & $89.50$ & $95.33$ \\
\bottomrule  
\end{tabular}  
\caption{Success rates ($\%$) of our method compared to Motion Planning Networks (MPNets)~\cite{Qureshi2019MotionPN} trained and evaluated on different datasets}
\label{tab:mpnets}
\end{table*}  
\paragraph{Neural Motion Planning} 
Motion Planning Networks (MPNets)~\citep{Qureshi2019MotionPN} proposed a similar method for neural motion planning, but there are a few key differences in both problem setup and system architecture. MPNets requires a ground-truth collision checker to connect sparse waypoints, plans in configuration space, and is not reactive to changing conditions. In the architecture, MPNets uses a trained neural sampler within a hierarchical bidirectional planner. The neural sampler is a fully-connected network that accepts the start, goal, and a flattened representation of the obstacle points as inputs and outputs a sample. MPNets guarantees completeness by using a traditional planner as a fallback if the neural sampler fails to produce a valid plan.

In addition to our data, we generated a set of tabletop problems, which we call \textit{MPNets-Style}, akin to the Baxter experiments in \citep{Qureshi2019MotionPN}, in order to fairly compare the two methods. The results of this experiment can be seen in Table \ref{tab:mpnets}. \mpinet requires a large dataset that covers the space of test problems to achieve compelling performance, while MPNets' utilization of a traditional planning system is much more effective with a small dataset or out of distribution problems. However, the MPNets architecture does not scale to more complex scenes, even with more data, as we show in Fig. \ref{fig:data-size}. When trained and evaluated on the Hybrid Planner-solvable dataset, MPNets succeeds in 
$65.28\%$ 
of the test set, whereas \mpinet succeeds in $95.33\%$, thus decreasing the failure rate by $7$X. Furthermore, as we show in Table \ref{tab:success-rates}, using the MPNets neural sampler trained with the \textit{Hybrid Planner} performs similarly to a uniform random sampler when both are embedded within the bidirectional MPNets planner.

\begin{table*}\centering\footnotesize
\begin{tabular}{@{}lccccccc@{}}
\toprule
& & & & \multicolumn{4}{c}{$\%$ Within} \\
\cmidrule(lr){5-8} 
& {$\%$ Env. Coll.} & {$\%$ Self Coll.} & {$\%$ Jnt Viol.} & {1\si{cm}} & {5\si{cm}} & {\ang{15}} & {\ang{30}}  \\
\midrule  
G. Fabrics \cite{VanWyk2022GeometricFG} & $8.61$ & $0.11$ & $0.44$ & $69.89$ & $75.17$ & $83.44$ & $85.11$  \\
STORM \cite{Bhardwaj2021FastJS} & $0.93$ & $0.11$ & $0.25$ & $79.81$ & $83.54$ & $81.57$ & $85.41$  \\
\addlinespace
\mpinet (Ours) \\  
\;\;\textsc{\textit{Hybrid Expert}} & $0.94$ &   $0.00$ & $0.00$ & $98.94$ & $99.72$ & $98.22$ & $99.00$ \\ 
\;\;\textsc{\textit{Global Expert}} & $13.78$  & $0.06$ & $0.00$ & $98.67$ & $99.89$ & $97.56$ & $99.11$ \\ 
\bottomrule  
\end{tabular}  
\caption{Failure Modes on problems solvable by both the global and hybrid planners}  
\label{tab:both-failure-modes}
\end{table*}  
\paragraph{Local Task Space Controllers}
Unlike planners, which succeed or fail in binary fashion, local policies will produce individual actions that, when rolled out, may fail for various reasons. We break down the various failure modes across the set of problems solvable by both experts in Table \ref{tab:both-failure-modes}. 

STORM~\cite{Bhardwaj2021FastJS} and Geometric Fabrics~\citep{VanWyk2022GeometricFG} make local decisions that can lead them to diverge from the target in complex scenarios, such as cluttered environments or those with pockets. For example, both STORM and Geometric Fabrics struggle to retract from a drawer and then reach into another drawer in a single motion without intermediate waypoints. While STORM, Geometric Fabrics, and \mpinet are all local policies, STORM and Geometric Fabrics rely on human tuning to achieve strong performance. Prior environment knowledge alongside expert tuning can lead to phenomenal results, but these parameter values do not generalize. We used a single set of parameters across all test environments just as we used a single set of weights for \mpinet. \mpinet encodes long-term planning information across a wide variety of environments, which makes it less prone to local minima, especially in unseen environments.

On problems solvable by the \textit{Hybrid Planner}, \mpinet ties or outperforms these other methods across nearly all metrics (see Table \ref{tab:hybrid-failure-modes}). On the set of problems solvable by the \textit{Global Planner}, \mpinet target convergence rate is consistently higher, while its collision rate ($11\%$) is worse than either STORM ($1.94\%$) or Geometric Fabrics ($7.83\%$) (see Table \ref{tab:global-failure-modes}).
Deteriorating performance on out-of-distribution problems is a typical downside of a supervised learning approach such as \mpinet. However, this could be improved with a more robust expert, \eg one with the consistency of our \textit{Hybrid Planner} but the success rate of the \textit{Global Planner}, with finetuning, or with DAgger \cite{Ross2011ARO}.

\subsection{Importance of the Expert Pipeline}
We observed that the choice of the expert pipeline affects the performance of \mpinet. We trained three policies: \mpinetg with $6.54$M demonstrations from the \textit{Global Planner}, \mpineth with $3.27$M demonstrations from the \textit{Hybrid Planner}, and \mpinetc with $3.27$M demonstrations from each. \mpinetc did not exhibit improved performance over either \mpineth or \mpinetg (see Appendix \ref{app:additional-experiments} for discussion). When evaluated on a test set of problems solvable by the \textit{Global Planner}, \mpinetg shows far better target convergence ($97.94\%$ vs. $87.72\%$) compared to \mpineth but worse obstacle avoidance ($21.94\%$ collision rate vs. $11\%$). Nonetheless, \mpineth is significantly better across all metrics when evaluated on problems solved by both experts as shown in Table \ref{tab:both-failure-modes}. We hypothesize that an expert combining the properties of these two--the consistency of the \textit{Hybrid Planner} and the generality of the \textit{Global Planner}, would further improve \mpinet's performance. We refer to \mpineth as \mpinet throughout the rest of the paper. 

\subsection{Comparison to Methods With Partial Observations}
\label{sec:partial-point-cloud}
In addition to demonstrating \mpinet' performance on a real robot system, we also compared \mpinet to the \textit{Global Planner} (AIT*~\citep{Strub2020AdaptivelyIT}) in a single-view depth camera setting in simulation. We evaluated on the test set of problems solvable by both the \textit{Global} and \textit{Hybrid Planners}. \mpinet only has a minor drop in success rate when using a partial point cloud vs. a full point cloud-- from $95.06\%$ to $93.22\%$ though the collision rate increases from $0.94\%$ to $3.06\%$ due to occlusions. For this experiment, we compared to the AIT* component of our \textit{Global Planner} alone to minimize false-positive solutions caused by the smoother's discrete collision checker (see discussion in Section \ref{subsec:data-plans}). We used a voxel-based reconstruction akin to the standard perception pipeline packaged with MoveIt~\cite{chitta2012moveit}. In our implementation, a voxel is filled only if a 3D point is registered within it. On the same test set using the voxel representation, AIT* produces plans with collisions on $16.41\%$ of problems. In this setting, \mpinet's collision rate is over $5X$ smaller than that of the \textit{Global Planner}.

\subsection{Ablations}
\label{subsec:ablations}

\begin{wrapfigure}{R}{0.45\textwidth}
  \vspace{-40pt}
  \begin{center}
    \includegraphics[width=0.4\textwidth]{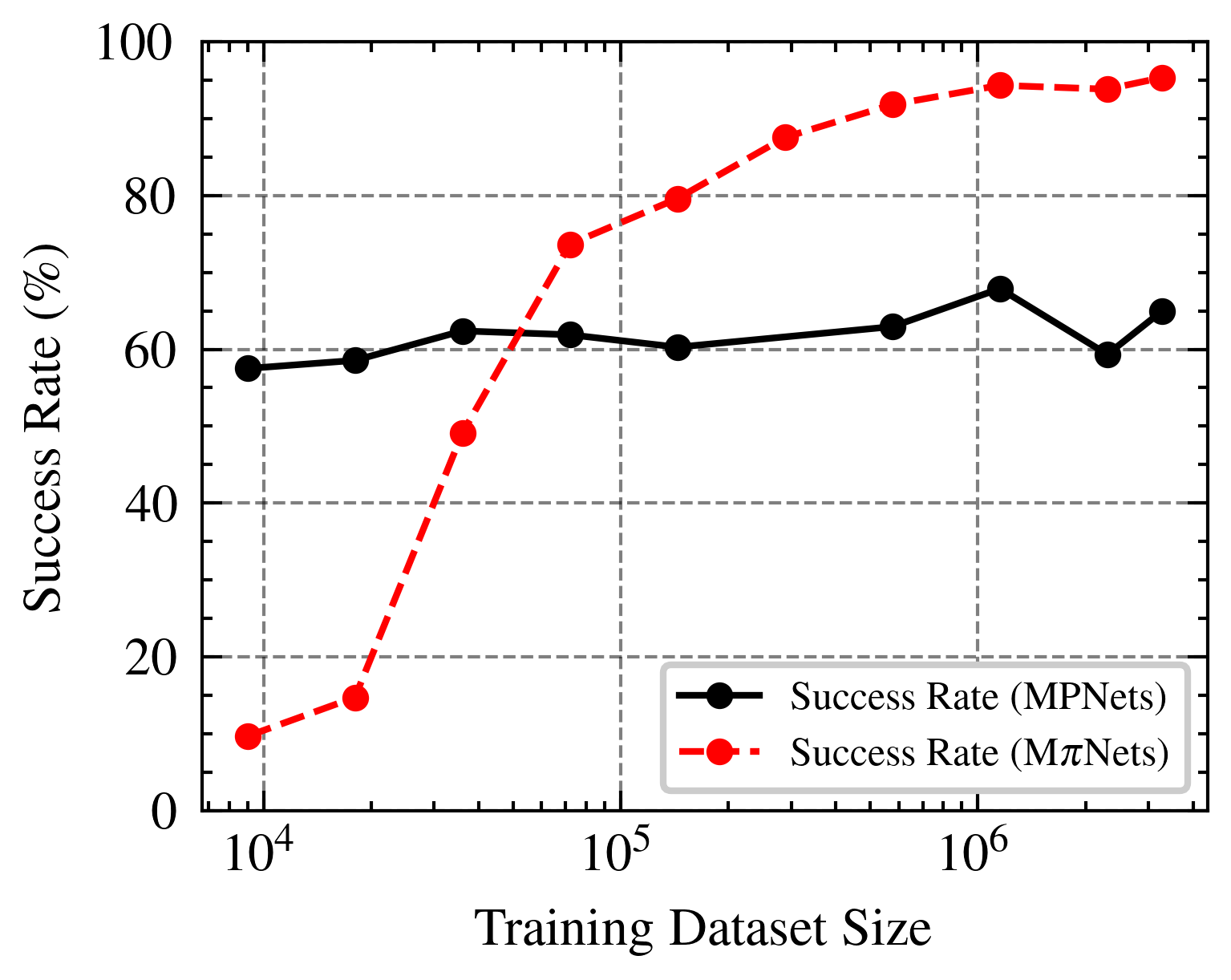}
  \end{center}
  \vspace{-10pt}
  \caption{\mpinet performance continues to increase with more training data, while MPNets performance stays relatively constant}
  \label{fig:data-size}
  \vspace{-10pt}
\end{wrapfigure}

We perform several ablations to justify our design decisions. All ablations were trained using the \textit{Hybrid Planner} dataset and evaluated on the \textit{Hybrid Planner}-solvable test set. More ablations and details can be found in Appendix \ref{app:additional-experiments}.

\textbf{\mpinet Performance Scales with More Data}
As shown in Fig. \ref{fig:data-size}, the performance of \mpinet continues to improve with more data, although it saturates at $1.1$M. Meanwhile, MPNets \cite{Qureshi2019MotionPN} has constant performance, demonstrating that our architecture is better able to scale with the data. 

\textbf{Robot Point Representation Improves Performance} 
Instead of representing the robot by its configuration vector, we insert the robot point cloud at the specific configuration. Without this representation, the success rate decreases from $95.33\%$ to $65.06\%$.

\textbf{Hindsight Goal Revision Improves Convergence} 
When trained without \textit{HGR}, \ie with the planner's original target given to the network, we see $58.11\%$ success rate vs. $95.33\%$ when trained with \textit{HGR}. In particular, only $60.28\%$ of trajectories get within 1cm of the target during evaluation.

\textbf{Noise Injection Improves Robustness}
When we train \mpinet without injecting noise into the input $q_t$, the policy performance decreases by $10.72\%$. 
\subsection{Dynamic Environments}
\mpinet is an instantaneous policy that assumes a static world at the time of inference. If the scene changes between inference steps, the policy will react accordingly. If the environment is continually changing--as is often the case in dynamic settings--\mpinet implicitly approximates the dynamic movement as a sequence of static motions. When the scene changes are slow, this assumption works well. When the changes are fast, it does not. To demonstrate this, we evaluated \mpinet in a static tabletop environment with a single, moving block placed on the table. We generated \num{1000} planning problems across the table with the block placed at different locations. We specifically chose problems where \mpinet succeeds when the block is stationary. When moving, the block follows a periodic curve in x and y, but the two curves have indivisible periods, preventing repetitive movement. We then moved the block at three different speeds: slow, medium, and fast and measured the success rate. At these speeds, \mpinet succeeds \num{88.1}\%, \num{57.4}\%, and \num{28.3}\% respectively.

\subsection{Real Robot Evaluation}
\label{sec:real-robot}
We deployed \mpinet on a $7$-DOF Franka Emika Panda robot with an extrinsically calibrated Intel Realsense L515 RGB-D camera mounted next to it. 
Depth measurements belonging to the robot are removed and re-inserted using a 3D model of the robot before inference with \mpinet. We created qualitative open-loop demonstrations in static environments and closed-loop demonstrations in dynamic ones. Rollouts are between $2$ and $80$ time steps long depending on the control loop frequency. See Appendix \ref{app:real-world-experiments} for system details.
Results can be viewed at \href{https://mpinets.github.io}{https://mpinets.github.io} and the attached video.
As can be seen, \mpinet~can achieve \textit{sim2real} transfer on noisy real-world point clouds in unknown and changing scenes.
\section{Limitations}
While \mpinet can handle a large class of problems, they are ultimately limited by the quality of the expert supervisor \red{and its need for a large, diverse dataset of training examples}. 
\red{Both generating the data and training \mpinet is computationally intensive, requiring access to equipment that is both economically and environmentally expensive.}
It will also struggle to generalize to out-of-distribution settings typical of any supervised learning approach. 
\red{When used on a real robot, performance will degrade as the robot's physical environment drifts from the training distribution. Likewise, performance will degrade with increasing point cloud noise.} In future work we aim to improve \mpinet with DAgger \cite{Ross2011ARO} or domain adaptation. In order to further enable safe operation in real robot systems, \mpinet could also be combined with a ground-truth or learnt collision checker such as SceneCollisionNet~\cite{Danielczuk2021ObjectRU}. \red{In future work, we intend to investigate how to incorporate a learned safety component to detect out-of-distribution input data and prevent unsafe operation.}

\section{Conclusion} 
\label{sec:conclusion}
\mpinet is a class of end-to-end neural policy policies that learn to navigate to pose targets in task space while avoiding obstacles. \mpinet show robust, reactive performance on a real robot system using data from a single, static depth camera. We train \mpinet with what is, as far as we are aware, the largest existing dataset of end-to-end motion for a robotic manipulator. Our experiments show that when applied to appropriate problems, \mpinet are significantly faster than a global motion planner and more capable than prior neural planners and manually designed local control policies. Code and data are publicly available at~\href{https://mpinets.github.io}{https://mpinets.github.io}.

\clearpage
\acknowledgments{We would like to thank the many people who have assisted in this research. In particular, we would like to thank Mike Skolones for supporting this research at NVIDIA, Balakumar Sundaralingam and Karl Van Wyk for their help in evaluating \mpinet and benchmarking it against STORM and Geometric Fabrics respectively; Ankur Handa, Chris Xie, Arsalan Mousavian, Daniel Gordon, and Aaron Walsman for their ideas on network architecture, 3D machine learning, and training; Nathan Ratliff and Chris Paxton for their help in shaping the idea early-on; Aditya Vamsikrishna, Rosario Scalise, Brian Hou, and Shohin Mukherjee for their help in exploring ideas for the expert pipeline, Jonathan Tremblay for his visualization expertise, Yu-Wei Chao and Yuxiang Yang for their help with using the Pybullet simulator, and Jennifer Mayer for editing the final paper.}

\bibliography{references}

\clearpage
\appendix
\section*{Appendix}
\section{Failure Modes Across All Test Sets}
\label{app:failure-modes}
In the main paper, we presented the breakdown of the failure modes  on the set of problems solvable by both the global and hybrid planners. In this section we present the failure modes separately across the two test sets. The \textit{Global Planner}-solvable test set is consistently the hardest for all methods, having the highest collision rates and target error. While STORM and Fabrics both see significant increases in target error, the change in collision rate is minor. When trained with the \textit{Global Expert}, \mpinet has the highest collision rate across all test sets, yet it also has the most consistent rollout accuracy. We attribute the collision rate to inconsistency in the \textit{Global Planner's} motion and the rollout accuracy to the high coverage of the problem space. When evaluated on the \textit{Global Planner}-solvable test set, \mpinet trained with the \textit{Hybrid Expert} also has its highest collision rate. We attribute this to distribution shift in the problem space. 
\begin{table*}[h]\centering\footnotesize
\begin{tabular}{@{}lccccccc@{}}
\toprule
& & & & \multicolumn{4}{c}{$\%$ Within} \\
\cmidrule(lr){5-8} 
& {$\%$ Env. Coll.} & {$\%$ Self Coll.} & {$\%$ Jnt Viol.} & {1\si{cm}} & {5\si{cm}} & {\ang{15}} & {\ang{30}}  \\
\midrule  
G. Fabrics \cite{VanWyk2022GeometricFG} & $8.17$ & $0.00$ & $0.39$ & $68.56$ & $73.33$ & $82.06$ & $84.00$  \\
STORM \cite{Bhardwaj2021FastJS} & $0.39$ & $0.11$ & $0.28$ & $83.11$ & $85.33$ & $90.00$ & $91.61$  \\
\addlinespace
\mpinet (Ours) \\  
\;\;\textsc{\textit{Hybrid Expert}} & $0.89$ & $0.00$ & $0.00$ & $98.83$ & $99.61$ & $98.83$ & $99.28$ \\ 
\;\;\textsc{\textit{Global Expert}} & $15.94$ & $0.00$ & $0.00$ & $99.00$ & $99.83$ & $97.06$ & $99.28$ \\ 
\bottomrule  
\end{tabular}  
\caption{Failure Modes on problems solvable by the hybrid planner}  
\label{tab:hybrid-failure-modes}
\end{table*}   
\begin{table*}[h]\centering\footnotesize
\begin{tabular}{@{}lccccccc@{}}
\toprule
& & & & \multicolumn{4}{c}{$\%$ Within} \\
\cmidrule(lr){5-8} 
& {$\%$ Env. Coll.} & {$\%$ Self Coll.} & {$\%$ Jnt Viol.} & {1\si{cm}} & {5\si{cm}} & {\ang{15}} & {\ang{30}}  \\
\midrule  
G. Fabrics \cite{VanWyk2022GeometricFG} & $7.83$ & $0.50$ & $0.33$ & $45.67$ & $57.33$ & $74.39$ & $78.22$  \\
STORM \cite{Bhardwaj2021FastJS} & $1.94$ & $0.11$ & $0.28$ & $71.33$ & $78.22$ & $64.44$ & $72.67$   \\
\addlinespace
\mpinet (Ours) \\  
\;\;\textsc{\textit{Hybrid Expert}} & $11.00$ & $0.78$ & $0.00$ & $87.72$ & $93.17$ & $84.56$ & $88.56$   \\ 
\;\;\textsc{\textit{Global Expert}} & $21.94$ & $0.00$ & $0.00$ & $97.94$ & $99.50$ & $96.56$ & $99.22$  \\ 
\bottomrule  
\end{tabular}  
\caption{Failure Modes on problems solvable by the global planner}  
\label{tab:global-failure-modes}
\end{table*}

\section{Expert Pipelines}
\label{app:expert-pipelines}

We present more details of our planning pipeline in this section.

\textbf{Global Planner} is composed of widely used off-the-shelf components. We first use inverse kinematics to convert our task space goals to configuration space, followed by AIT*~\cite{Strub2020AdaptivelyIT} in configuration space, and finally, spline-based, collision-aware trajectory smoothing~\cite{Hauser2010FastSO}. We use IKFast~\cite{diankov_thesis} for inverse kinematics, OMPL~\cite{sucan2012the-open-motion-planning-library} for AIT*, and Pybullet Planning for the smoothing implementation~\cite{Garrett2018}. To manage the compute load when generating a large dataset of trajectories, we employed a time-out with AIT* of 20 seconds.

\textbf{Hybrid Expert} is designed to produce consistent motion in task space. We start by using AIT*~\citep{Strub2020AdaptivelyIT} with a 2 second timeout to plan for a floating end effector, \ie one not attached to a robot arm, and then use Geometric Fabrics \cite{VanWyk2022GeometricFG} to follow the path. Geometric Fabrics are deterministic and geometrically consistent~\cite{VanWyk2022GeometricFG} local controllers, but they struggle to solve the problems in our dataset without assistance from a global planner. Geometric Fabrics are highly local, and even with dense waypoints given by a global planner, they can run into local minima, which in turn generate trajectories with highly variable velocity. We use a combination of spline-based smoothing and downsampling~\cite{Hauser2013FastIA} to create a consistent configuration space velocity profile across our dataset. 

\textbf{Consistency}
We use the term \textit{consistency} to describe a qualitative characteristic of a planner and its learnability. Specifically, we use it to describe two quantities: 1) expert quality and 2) repeatability of the planner. 
\citet{robomimic2021} demonstrate how Imitation Learning performance varies depending on expert quality. Among the metrics they use to describe expert quality, they demonstrate the importance of expert trajectory length. \mpinet employs task-space goals, and the \textit{Hybrid Planner} produces shorter task space paths. Across our test dataset of global and hybrid solvable problems, the \textit{Hybrid Planner}'s end effector paths average 57cm ± 31cm and the total orientation distance traveled in 95° ± 52°. Meanwhile, the \textit{Global Planner}'s paths average 61cm ± 39cm and 113° ± 55°, respectively. 
 
Repeatable input-output datasets are important for deep learning systems. Prior works have shown the deep learning systems deteriorate or require more data when using noisy labels~\citep{Misra2016SeeingTT, Joulin2016LearningVF}. Both the \textit{Global Planner} and \textit{Hybrid Planner} are sampling-based planners and do not produce repeatable paths by their very nature. Yet, the \textit{Hybrid Planner} uses sampling to plan in a lower-dimensional state space---6D pose space---while the \textit{Global Planner} samples in 7D configuration space. We use a naive sampler, so the lower dimensionality of the \textit{Hybrid Planner's} sampler implies that it's typical convergence rate will be faster. After planning, the \textit{Hybrid Planner} employs Geometric Fabrics~\citep{VanWyk2022GeometricFG} to follow the task-space trajectory. Geometric Fabrics are deterministic, which further promotes repeatability in the final, configuration space trajectories. Meanwhile, the \textit{Global Planner} uses a randomized smoothing algorithm that is not deterministic. Taking these individual components together, we expect the \textit{Hybrid Planner's} solutions on similar problems to be typically more alike than the \textit{Global Planner's} solutions to the same problems.

\section{Network Architecture}
\label{app:network-architecture}
Our PointNet++ architecture has three set abstraction groups followed by three fully connected layers. The first set abstraction layer performs iterative furthest point sampling to construct a set of $512$ points, then it does a grouping query within $5\si{\cm}$ of at most $128$ points. Finally, there is a local PointNet~\cite{Qi2017PointNetDL} made up of layers of size $4$, $64$, $64$, $64$ respectively. The second set abstraction is lower resolution, sampling $128$ furthest points and then grouping at most $128$ points within a $30\si{\cm}$ radius. The corresponding PointNet is made up of layers of size $64$, $128$, $128$, and $256$ respectively. Our third set abstraction layer skips the furthest point sampling, groups all points together, and uses a final PointNet with layers of size $256$, $512$, $512$, \num{1024} respectively. Finally, after the set abstraction layers, we have three fully connected layers with \num{4096}, \num{4096}, and \num{2048} dimensions respectively. In between these layers, we use group norm and Leaky ReLU.

The output of our point cloud encoder is a \num{2048} dimensional embedding.
The robot configuration encoder and the displacement decoder are both fully connected multilayer perceptrons with Leaky ReLU activation functions \cite{Maas13rectifiernonlinearities}. The robot configuration encoder maps our 7 dimensional input to a 64 dimensional output and has four hidden layers with $32$, $64$, $128$, and $128$ dimensions respectively. The displacement decoder maps the combined embeddings from the point cloud and robot configuration encoders, which together have \num{2112} dimensions, to the $7$ dimensional normalized displacement space. The decoder has three hidden layers with $512$, $256$, and $128$ dimensions respectively. Our entire architecture together has $19$ million parameters.

\section{Data Generation Pipeline}
\label{app:data-generation}
We used the same procedural data generation pipeline to generate data for training as well as inference test problems. We will be releasing the code to generate the data alongside our generated data sets upon publication.

\textbf{Tabletop} 
The dimensions of the table, including height, are randomized, as well as whether the table has an L-bend around the robot. The table itself is always axis-oriented. Table height ranges from \num{0} to \num{40}\si{cm}. Table edges are chosen independently, \eg the maximum x value for a table is chosen from a uniform distribution, and the center of the tables is not fixed. The front table can range between \num{90} and \num{110}\si{cm} deep and between \num{205} and \num{240}\si{cm} wide. When there is an L-bend, the side table ranges from \num{90} to \num{247.5}\si{cm} deep and \num{42.5} to \num{72.5}\si{cm} wide. After generating the table, a random assortment of boxes and cylinders are placed on the table facing upward, \ie cylinders are on their flat edge. There are between \num{3} and \num{15} objects in each scene. These objects are between \num{5} and \num{35}\si{cm} tall. The side dimensions of the boxes, as well as the radius of the cylinders, are between \num{5} and \num{15}\si{cm}.

\textbf{Cubby} 
The dimensions of the cubby, the wall-thickness, the number of cubbies, and orientation of the entire fixture are randomized. We start by constructing a two-by-two cubby and then modify it to randomize the number of cubby holes. The wall thickness is chosen to be between \num{1} and \num{2}\si{cm}. Similar to the tabletop, cubby edges are chosen independently, which implicitly set the center. The overall fixture is ranges from \num{120} to \num{160}\si{cm} wide, \num{20} to {35}\si{cm} deep, and between \num{30} and \num{60}\si{cm} tall. The horizontal and vertical center dividers are then placed randomly within a \num{20}\si{cm} range. Finally, we apply a random yaw rotation of up to \num{40}\degree around the fixture's central axis. For roughly half of the cubby environments, we modify the cubby to reduce the number of cubby holes. To do this, we select two random, collision-free robot configurations in two separate cubby holes and then merge the cubby holes necessary to create a collision-free path between them.

\textbf{Dresser} 
The dimensions of the dresser, the placement of the drawers, and the orientation of the entire fixture are randomized. The dresser side walls, drawer side walls, and drawer faces are always \num{1}, \num{1.9}, and {0.4}\si{cm} thick respectively. Unlike the other two environments, dimensions for the dresser are chosen randomly, as is the center point for the fixture. The dresser dimensions range from 80 to 120\si{cm} wide, 20 to 40\si{cm} deep, and 55 to 85\si{cm} tall. The dresser is always placed on the ground randomly in reachable space of the robot, with a random orientation around its central yaw axis. We next construct the drawers. We randomly choose a direction in which to split the dresser and then split it into two drawers. We perform this recursively within each drawer, stopping according to a decaying probability function. Finally, we open two drawers within reachable space.

\textbf{Initial Configurations and Target Poses}
After generating a random fixture, we search for valid start and goal configurations. We first look for target poses with reasonable orientations--in a grasping pose for the tabletop, pointing approximately inward for a cubby, or pointing approximately downward in a drawer. We choose pairs of these targets, solve for a collision-free inverse kinematics solution for each target, and consider these configuration space solutions to be candidates for the start or end of a trajectory. We also add a set of collision-free neutral configurations to the mix. These neutral configurations are generated by adding uniform randomness to a seed neutral configuration. From this set of task-space targets and corresponding collision-free configuration space solutions, we select pairs to represent a single planning problem. For each pair selected, we use the \textit{Global Planner} to verify that a smooth, collision-free planning solution exists.

\section{Training \mpinet}
\label{app:training-mpinet}
We implemented \mpinet in PyTorch and used the Adam optimizer with a learning rate of 0.0004. We trained it across 8 NVIDIA Tesla V100 GPUs for a week.
\section{Inference with \mpinet}
We used separate inference hardware for our simulated experiments and the hardware demonstrations For our simulated experiments, we use a desktop with CPU Intel(R) Core(TM) i9-9820X CPU @ 3.30GHz, GPU NVIDIA A6000, and 64GB of RAM. For our hardware demonstrations, we used a desktop with CPU Intel(R) Core(TM) i7-7800X CPU @ 3.50GHz, GPU NVIDIA Titan RTX, and 32GB of RAM.

\section{Quantitative Metrics}
\label{app:metrics}
\paragraph{Success Rate} A trajectory is considered a success if the rollout position and orientation target errors are below~\SI{1}{\cm} and \SI{15}{\degree} respectively and there are no physical violations. To avoid erroneously passing a trajectory that ends on the wrong side of a narrow structure, we also ensure that the end effector is within the correct final volume and likewise avoids incorrect volumes. For the cubby and dresser environments, these volumes are individual cubbies or drawers. 

\paragraph{Time} After setting up each planning problem, we measured the wall time for each \textit{successful} trajectory. We also measure \textit{Cold Start (CS) Time}, the average time to react to a new planning problem. While both expert pipelines have to compute the entire path, the local controllers only need time to compute a single action. We only consider the cold-start time here, but if the new planning problem is sufficiently similar to a previous one--such as a minor change in the environment or target--a global planning system could employ an optimizer that can replan quickly~\cite{Mukadam-IJRR-18}. 

\paragraph{Rollout Target Error} We calculate both position and orientation errors from the target for the final end effector pose in the trajectory. We measure position error with Euclidean distance and orientation error with the metric described by \citet{Wunsch1997RealtimePE}.

\paragraph{Collisions} A trajectory can have two types of fatal collisions--when the robot collides with itself or when the robot collides with the scene. When checking for collisions, we use an ensemble of collision checkers to ensure fairness. Collision checking varies across algorithmic implementations, \eg our AIT* implementation uses meshes to check scene collisions, while STORM~\cite{Bhardwaj2021FastJS} and Geometric Fabrics~\cite{VanWyk2022GeometricFG} use a sphere-based approximation of the robot's geometry. A trajectory is only considered to be in collision if the entire ensemble agrees.

\paragraph{Smoothness} We use Spectral Arc Length~(SPARC)~\cite{Balasubramanian2015OnTA} to measure smoothness. \citet{Balasubramanian2015OnTA} use a SPARC threshold of \SI{-1.6}{} as sufficiently smooth for reaching tasks. This measurement qualitatively describes the behavior of our benchmark algorithms well, so we used the same threshold for sufficiency. We therefore consider a path to be smooth if both its joint-space trajectory and end effector trajectory have SPARC values below \SI{-1.6}{}.

\section{Local Policy Implementations}
\label{app:local-policies}
Both STORM~\cite{Bhardwaj2021FastJS} and Geometric Fabrics~\cite{VanWyk2022GeometricFG} require expert tuning to achieve compelling performance, and we worked closely with the authors of these papers to tune them as best as possible for our evaluation. We train a single network on all three environment types, so similarly use a single set of tuning parameters for each algorithm over the entire evaluation set. 

\section{MPNets Implementation and Data}
\label{app:mpnets}

In the original paper, \citet{Qureshi2019MotionPN} trained MPNets for execution on the Baxter robot using a dataset of \num{10} different tabletop environments, each with 900 plans. Then, it was evaluated in the same environments using 100 unseen start and goal configurations in each. In total, their real-robot dataset was \num{10000} problems.

To compare fairly to MPNets, we generated an analogous set of \num{10000} problems within \num{10} tabletop environments, which we call the \textit{MPNets-Style} dataset. We re-implemented the MPNets-algorithm based on their open source implementation at 
\href{https://github.com/anthonysimeonov/baxter_mpnet_experiments}{https://github.com/anthonysimeonov/baxter\_mpnet\_experiments}. 

After we trained our implementation of their model on the MPNets-Style data, it achieved a similar success rate as the one quoted in their paper for the Baxter experiments ($78\%$ vs. $85\%$). We attribute the performance difference to the increased complexity of our environments, which, unlike the original dataset, have varying table geometry in addition to object placement. In the original paper, they quote planning as taking $1$ second on average. Our re-implementation took $2.47$ seconds on average with a median of $0.02$ seconds. Again, we attribute this difference to the increased complexity, given that the median time is so far below the mean. Just as they do in the open source implementation, we employ hierarchical re-planning, but we do not fall back to a traditional planner. If given access to a collision checker, both \mpinet and MPNets can use a similar fallback to re-plan, thus achieving theoretically complete performance.   

We used the same training setup described in Appendix \ref{app:training-mpinet} to train MPNets. When trained on the \mpinet data set, \ie \num{3.27}M demonstrations from the\textit{Hybrid Planner}, MPNets converged within 15 hours.

\section{Additional Experiments}
\label{app:additional-experiments}

\begin{wrapfigure}{R}{0.45\textwidth}
  \vspace{-40pt}
  \begin{center}
    \includegraphics[width=0.4\textwidth]{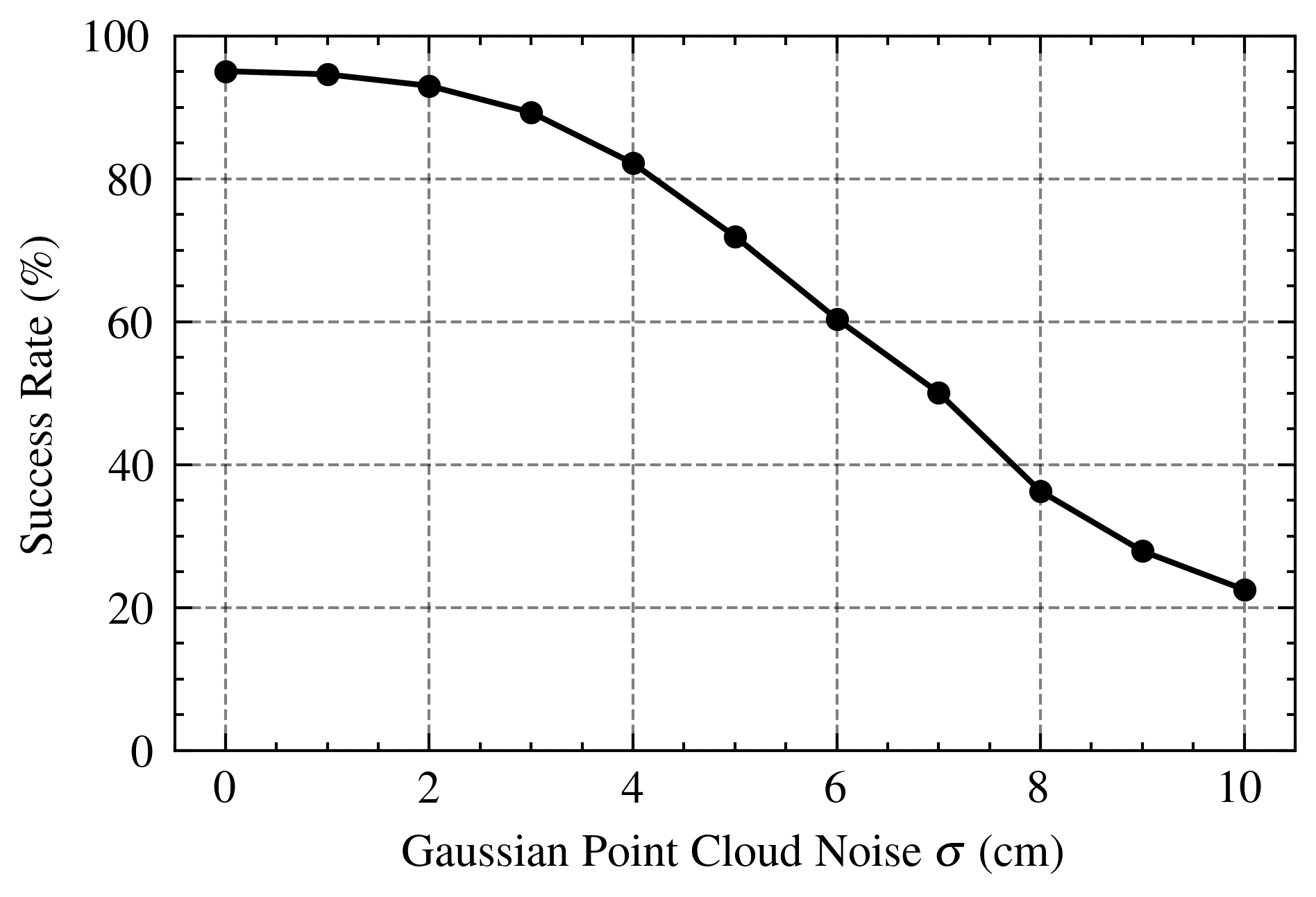}
  \end{center}
  \vspace{-10pt}
  \caption{After injecting Gaussian noise into the point clouds, \mpinet performance stays fairly constant up until $\sigma=3\si{\cm}$ when success rate is \num{89.28}\%.}
  \label{fig:pc-noise}
  \vspace{-10pt}
\end{wrapfigure}

\textbf{Training with Mean Squared Error Loss Increases Collisions}\label{par:mse}
When trained with a loss of mean-squared-error in configuration space, \mpinet has a similar success rate--$94.56\%$ vs. $95.33\%$--but scene collision rate is significantly higher at $2.39\%$ vs $0.89\%$.

\textbf{Representing the Target in Point Cloud Improves Performance} When trained with the target fed explicitly through a separate MLP encoder as a position and quaternion, \mpinet succeeds less--$88.83\%$ vs. $95.33\%$ when the target is specified within the point cloud. In particular, only $91.61\%$ of trajectories get within 1cm of the target vs. $98.83\%$ with the point cloud-based target.

\textbf{Training with Collision Loss Improves Collision Rate} When trained without the collision loss, \mpinet collides more often--$2.11\%$ vs $0.89\%$ when trained with the collision loss. 

\textbf{Training with the Configuration Encoder Improves Success Rate} When trained with no robot configuration encoder, \ie with only the point cloud encoder, \mpinet has success rate of $94.17\%$ vs $95.33\%$ when trained with both encoders.

\textbf{\mpinet is Robust to Point Cloud Noise Up to \num{3.2}\si{cm}} Figure \ref{fig:pc-noise} shows \mpinet success rate on the set of problems solvable by both planners when random Gaussian noise is added to the point cloud. Model performance stays above 90\% until noise reaches \num{3}\si{cm} at which point success drops to \num{89.28}\%.

\textbf{\mpinet is Robust to Varying Point Cloud Shapes} To evaluate performance in out-of-distribution geometries, we replaced all tabletop objects in test set of problems solvable by the \textit{Hybrid Planner} with randomly meshes from the YCB dataset \citep{Calli2015TheYO}. For each tabletop primitive, we sampled a mesh from the dataset and transformed it so that the bounding boxes of the primitive and mesh were aligned and of identical size. Note that in these modified scenes, the primitives-based \textit{Hybrid Planner} solution is still valid. \mpinet succeeded in {88.33}\% in this YCB-tabletop test set, whereas with the original primitives, it succeeds in \num{94.67}\%. Note that the network was not trained with these geometries---we would expect even higher performance if these meshes were included in the training set.

\textbf{\mpinet is Not Suitable for Unsolvable Problems} To evaluate performance on unsolvable problems, we generated a set of 800 planning problems in randomized tabletops where the target is in collision with the table or an object on the table. When used for these problems, \mpinet showed a \num{64.25}\% collision rate.

\textbf{\mpinet is Not Improved by Combining Experts} We trained \mpinetc on a combination of \num{3.27}M demonstrations each from the \textit{Hybrid Planner} and \textit{Global Planner}. Environments may have overlapped in these data sets, but entire problems, \ie environment, start, and goal, did not. In problems solvable by the global planner, \mpinetc---like \mpinetg---outperformed \mpineth in terms of target convergence (\num{97.17}\% vs \num{87.72}\%). While its collision rate is lower than \mpinetg, (\num{18.56}\% vs \num{21.94}\%) \mpinetc's collision rate is still significantly higher than \mpineth (\num{11}\%). The behavior of \mpinetc is essentially an average of \mpinetg and \mpineth, which we attribute to the lack of easily learnable obstacle avoidance behavior by the \textit{Global Planner}.  These demonstrations equate to additional noise in the training data, which creates less successful obstacle avoidance behavior. In future work, we intend to explore how to robustly combine experts for improved performance.

\section{Real-World Experiments}
\label{app:real-world-experiments}

We demonstrated \mpinet in a variety of table top problems using a Franka Emika Panda $7$-DOF manipulator. A calibrated Intel Realsense L515 RGB-D camera is placed in front of the robot's workspace, viewing the table and potential obstacles on top of it. Point cloud measurements are filtered to remove all points belonging to the robot geometry. The remaining cloud is downsampled to 4096~points and treated as the obstacle. The filtering process runs at \SI{9}{\Hz}. We investigated two different control methods:

\paragraph{Open-Loop Motion:}
Using a fixed, user-defined goal location and the current depth observation, \mpinet is rolled out over 80~timesteps or until goal convergence.
The resulting path is used to compute a time-parametrized trajectory~\cite{kunz2012time} which is then tracked by a position controller.
The videos listed under ``Open Loop Demonstrations'' at \href{https://mpinets.github.io}{https://mpinets.github.io} show a mix of sequential motions toward pre-defined goals. In some of the examples, the objects are static throughout the video and in others, we re-arrange the objects throughout the video.
Despite the changing scene, these are still open-loop demos. While the motions adapt to changing obstacles in the scene, the policy only considers scene changes that happen before execution of a trajectory. This is because the point cloud observations are only updated once the robot reaches its previous target.

\paragraph{Closed-Loop Motion:}
To account for dynamic obstacles \mpinet is rolled out for a single timestep at the same frequency as the point cloud filter operates~(\SI{9}{\Hz}). A time-parametrized trajectory is generated by linearly interpolating $\approx70\%$ of the rolled out path. As in the open-loop case the resulting trajectory is tracked by a PD controller controller at \SI{1}{\kHz}. The videos listed under ``Closed Loop, Dynamic Scene Demonstrations'' at \href{https://mpinets.github.io}{https://mpinets.github.io} show examples of boxes thrown into the robot's path while it is moving towards a user-defined target. The evasive maneuver shows \mpinet' ability to react to dynamic obstacles.

\section{Limitations}
\label{app:limitations}

\paragraph{Training Distribution} The limitations of the \textit{Hybrid Planner} translate to limitations in the trained policy network. Certain target poses and starting configurations can create unanticipated behavior. When target poses are narrowly out of distribution, the rollout fails to converge to the target, but as a target poses drifts further from the training distribution, behavior becomes erratic. Likewise, random, initial configurations--such as from rejection-sampling based inverse kinematics--can create unexpected behavior, but we did not observe this in our real robot trials running the policy continuously to a sequence of points. With an improved expert, \eg one with the consistency of our \textit{Hybrid Expert} and guaranteed convergence of the \textit{Global Planner}, we anticipate that the occurrence of failure cases will diminish. We also do not expect the network to generalize to wholly unseen geometries without more training data. But, in future work, we aim to improve the generalization of this method, much in the way that Large Language Models \cite{CLIP2021} continue to improve generalization through data. 

\paragraph{Real Robot System} In order to ensure safe operation in a real-robot system, \mpinet could be combined with a collision checker--either one with ground-truth or a learned, such as Scene Collision Net~\cite{Danielczuk2021ObjectRU}. The collision checker could be used to a) stop the robot before hitting collisions b) make small perturbations to nudge the policy back into distribution or c) enable a traditional planner to plan to the goal. In a physical system, not all problems will have feasible solutions. As discussed in Appendix \ref{app:additional-experiments}, \mpinet will often collide in these scenarios, underscoring the need for some additional safety mechanisms to prevent catastrophic behavior. Additionally, \mpinet has no concept of history and can collide with the scene if, for example, the robot arm blocks the camera mid-trajectory. To mitigate this, the perception system could employ a historical buffer or filter to maintain some memory of the scene.

\paragraph{Emergent Behavior} In some of our test problems, we observed that \mpinet produces a rollout where the final gripper orientation is $180\si{\degree}$ off from the target about the gripper's central axis (\ie the central axis parallel to the fingers). In the test set of problems solvable by the \textit{Global Planner}, this occurs in $2.44\%$ of rollouts. We suspect this behavior is due to the near-symmetry in the gripper's mesh about this axis. The minor differences between the two sides of the gripper may not provide enough information for the Pointnet++ encoder to distinguish between these two orientations. While the rollout does not match the requested problem, this behavior can be desirable in some circumstances. For example, because grasps are symmetric with the Franka Panda gripper, a $180\si{\degree}$ rotation is preferable if it reduces the likelihood of a collision. For applications where this behavior is unacceptable, we could replace the target representation in the point cloud with points sampled from a mesh with no symmetry.

\section{Experimental Results per Environment}
\label{app:experimental-results-per-environment}
In this section, we present the evaluation metrics broken down by environment type. However, we omit Cold Start Time because for global methods, it is the same as the total time and for local methods, the type of environment does not affect startup or reaction time. 

The Tabletop environment is the least challenging with the highest success rates for all methods. In general, the dresser environment is the most challenging due to its complex geometry, as evidenced by the high collision rates. When trained with the \textit{Hybrid Expert}, \mpinet has the highest rollout target error in the cubby problems solvable by \textit{Global Planner}. Since \mpinet trained with the \textit{Global Expert} does not have this issue, we attribute it to a lack of adequate coverage in the training dataset.

\begin{table*}\centering\footnotesize
\addtolength{\tabcolsep}{-2pt}
\begin{tabular}{@{}lccccc@{}}
\toprule
& & \multicolumn{3}{c}{Success Rate ($\%$)} \\
\cmidrule(lr){3-5}
& Soln. Time (\si{\second}) & Global & Hybrid & Both & Smooth ($\%$) \\
\midrule
\addlinespace
Global Planner \cite{Strub2020AdaptivelyIT} & $16.56 \pm 0.88$ & $100$ & $73.50$ & $100$ & $56.86$ \\
Hybrid Planner & $6.82 \pm 1.50$ & $44.33$ & $100$ & $100$ & $99.22$ \\
\addlinespace
G. Fabrics \cite{VanWyk2022GeometricFG} & $0.11 \pm 0.06$ & $37.83$ & $66.67$ & $65.83$ & $88.61$ \\
STORM \cite{Bhardwaj2021FastJS} & $3.65 \pm 1.64$ & $53.50$ & $76.67$ & $77.33$ & $59.72$ \\
\addlinespace
MPNets~\cite{Qureshi2019MotionPN} \\
\;\;\textit{Hybrid Expert} & $2.68 \pm 17.39$ & $44.67$ & $59.00$ & $66.17$ & $17.26$ \\
\;\;\textit{Random} & $0.06 \pm 0.06$ & $32.17$ & $50.17$ & $53.67$ & $100.00$ \\
\addlinespace
\mpinet (Ours) \\
\;\;\textsc{\textit{Hybrid Expert}} & $0.33 \pm 0.08$ & $67.00$ & $94.33$ & $93.17$ & $93.06$ \\
\;\;\textsc{\textit{Global Expert}} & $0.34 \pm 0.07$ & $74.83$ & $81.50$ & $80.00$ & $93.44$ \\
\bottomrule
\end{tabular}
\caption{Algorithm performance on cubby problems sets solvable by planner types. All prior methods use state-information and a oracle collision checker while \mpinet only needs a point cloud}
\label{tab:cubby-success-rates}
\addtolength{\tabcolsep}{2pt}
\end{table*}

\begin{table*}[h]\centering\footnotesize
\begin{tabular}{@{}lccccccc@{}}
\toprule
& & & & \multicolumn{4}{c}{$\%$ Within} \\
\cmidrule(lr){5-8}
& {$\%$ Env. Coll.} & {$\%$ Self Coll.} & {$\%$ Jnt Viol.} & {1\si{cm}} & {5\si{cm}} & {\ang{15}} & {\ang{30}}  \\
\midrule
G. Fabrics \cite{VanWyk2022GeometricFG} & $5.00$ & $0.17$ & $0.67$ & $40.17$ & $57.83$ & $84.67$ & $89.17$ \\
STORM \cite{Bhardwaj2021FastJS} & $0.50$ & $0.00$ & $0.50$ & $79.33$ & $85.33$ & $69.17$ & $80.33$ \\
\addlinespace
\mpinet (Ours) \\
\;\;\textsc{\textit{Hybrid Expert}} & $10.67$ & $0.17$ & $0.00$ & $75.83$ & $84.50$ & $75.83$ & $81.67$ \\
\;\;\textsc{\textit{Global Expert}} & $23.17$ & $0.00$ & $0.00$ & $99.17$ & $100.00$ & $99.33$ & $100.00$ \\
\bottomrule
\end{tabular}
\caption{Failure Modes on cubby problems solvable by the global planner}
\label{tab:cubby-global-failure-modes}
\end{table*}

\begin{table*}[h]\centering\footnotesize
\begin{tabular}{@{}lccccccc@{}}
\toprule
& & & & \multicolumn{4}{c}{$\%$ Within} \\
\cmidrule(lr){5-8}
& {$\%$ Env. Coll.} & {$\%$ Self Coll.} & {$\%$ Jnt Viol.} & {1\si{cm}} & {5\si{cm}} & {\ang{15}} & {\ang{30}}  \\
\midrule
G. Fabrics \cite{VanWyk2022GeometricFG} & $4.83$ & $0.00$ & $1.00$ & $72.50$ & $83.00$ & $95.83$ & $96.33$ \\
STORM \cite{Bhardwaj2021FastJS} & $0.17$ & $0.17$ & $0.33$ & $87.33$ & $89.33$ & $89.17$ & $91.67$ \\
\addlinespace
\mpinet (Ours) \\
\;\;\textsc{\textit{Hybrid Expert}} & $0.50$ & $0.00$ & $0.00$ & $99.83$ & $99.83$ & $100.00$ & $100.00$ \\
\;\;\textsc{\textit{Global Expert}} & $16.67$ & $0.00$ & $0.00$ & $99.50$ & $100.00$ & $99.83$ & $100.00$ \\
\bottomrule
\end{tabular}
\caption{Failure Modes on cubby problems solvable by the hybrid planner}
\label{tab:cubby-hybrid-failure-modes}
\end{table*}

\begin{table*}[h]\centering\footnotesize
\begin{tabular}{@{}lccccccc@{}}
\toprule
& & & & \multicolumn{4}{c}{$\%$ Within} \\
\cmidrule(lr){5-8}
& {$\%$ Env. Coll.} & {$\%$ Self Coll.} & {$\%$ Jnt Viol.} & {1\si{cm}} & {5\si{cm}} & {\ang{15}} & {\ang{30}}  \\
\midrule
G. Fabrics \cite{VanWyk2022GeometricFG} & $5.00$ & $0.00$ & $1.17$ & $72.33$ & $84.33$ & $96.33$ & $97.33$ \\
STORM \cite{Bhardwaj2021FastJS} & $0.00$ & $0.00$ & $0.00$ & $88.33$ & $89.00$ & $89.33$ & $91.67$ \\
\addlinespace
\mpinet (Ours) \\
\;\;\textsc{\textit{Hybrid Expert}} & $0.50$ & $0.00$ & $0.00$ & $99.83$ & $100.00$ & $99.83$ & $100.00$ \\
\;\;\textsc{\textit{Global Expert}} & $18.17$ & $0.00$ & $0.00$ & $99.00$ & $100.00$ & $100.00$ & $100.00$ \\
\bottomrule
\end{tabular}
\caption{Failure Modes on cubby problems solvable by both the global and hybrid planners}
\label{tab:cubby-both-failure-modes}
\end{table*}

\begin{table*}\centering\footnotesize
\addtolength{\tabcolsep}{-2pt}
\begin{tabular}{@{}lccccc@{}}
\toprule
& & \multicolumn{3}{c}{Success Rate ($\%$)} \\
\cmidrule(lr){3-5}
& Soln. Time (\si{\second}) & Global & Hybrid & Both & Smooth ($\%$) \\
\midrule
\addlinespace
Global Planner \cite{Strub2020AdaptivelyIT} & $16.97 \pm 0.81$ & $100$ & $66.83$ & $100$ & $75.63$ \\
Hybrid Planner & $9.19 \pm 2.81$ & $37.33$ & $100$ & $100$ & $99.82$ \\
\addlinespace
G. Fabrics \cite{VanWyk2022GeometricFG} & $0.26 \pm 0.12$ & $15.00$ & $25.83$ & $28.50$ & $78.94$ \\
STORM \cite{Bhardwaj2021FastJS} & $5.54 \pm 1.84$ & $24.17$ & $58.50$ & $62.00$ & $83.22$ \\
\addlinespace
MPNets~\cite{Qureshi2019MotionPN} \\
\;\;\textit{Hybrid Expert} & $15.55 \pm 46.31$ & $12.83$ & $41.83$ & $41.67$ & $26.68$ \\
\;\;\textit{Random} & $1.61 \pm 7.38$ & $8.33$ & $27.50$ & $31.17$ & $100.00$ \\
\addlinespace
\mpinet (Ours) \\
\;\;\textsc{\textit{Hybrid Expert}} & $0.34 \pm 0.06$ & $78.67$ & $97.00$ & $96.33$ & $91.56$ \\
\;\;\textsc{\textit{Global Expert}} & $0.33 \pm 0.05$ & $72.33$ & $77.33$ & $82.17$ & $94.89$ \\
\bottomrule
\end{tabular}
\caption{Algorithm performance on dresser problems sets solvable by planner types. All prior methods use state-information and a oracle collision checker while \mpinet only needs a point cloud}
\label{tab:dresser-success-rates}
\addtolength{\tabcolsep}{2pt}
\end{table*}

\begin{table*}[h]\centering\footnotesize
\begin{tabular}{@{}lccccccc@{}}
\toprule
& & & & \multicolumn{4}{c}{$\%$ Within} \\
\cmidrule(lr){5-8}
& {$\%$ Env. Coll.} & {$\%$ Self Coll.} & {$\%$ Jnt Viol.} & {1\si{cm}} & {5\si{cm}} & {\ang{15}} & {\ang{30}}  \\
\midrule
G. Fabrics \cite{VanWyk2022GeometricFG} & $17.17$ & $0.83$ & $0.17$ & $19.83$ & $26.33$ & $57.83$ & $62.33$ \\
STORM \cite{Bhardwaj2021FastJS} & $4.83$ & $0.17$ & $0.33$ & $42.67$ & $51.67$ & $45.17$ & $53.83$ \\
\addlinespace
\mpinet (Ours) \\
\;\;\textsc{\textit{Hybrid Expert}} & $17.00$ & $0.83$ & $0.00$ & $98.00$ & $98.67$ & $93.50$ & $94.33$ \\
\;\;\textsc{\textit{Global Expert}} & $26.67$ & $0.00$ & $0.00$ & $100.00$ & $100.00$ & $99.00$ & $99.83$ \\
\bottomrule
\end{tabular}
\caption{Failure Modes on dresser problems solvable by the global planner}
\label{tab:dresser-global-failure-modes}
\end{table*}

\begin{table*}[h]\centering\footnotesize
\begin{tabular}{@{}lccccccc@{}}
\toprule
& & & & \multicolumn{4}{c}{$\%$ Within} \\
\cmidrule(lr){5-8}
& {$\%$ Env. Coll.} & {$\%$ Self Coll.} & {$\%$ Jnt Viol.} & {1\si{cm}} & {5\si{cm}} & {\ang{15}} & {\ang{30}}  \\
\midrule
G. Fabrics \cite{VanWyk2022GeometricFG} & $18.33$ & $0.00$ & $0.17$ & $36.00$ & $39.00$ & $61.00$ & $66.00$ \\
STORM \cite{Bhardwaj2021FastJS} & $0.83$ & $0.00$ & $0.33$ & $65.33$ & $67.67$ & $90.17$ & $91.00$ \\
\addlinespace
\mpinet (Ours) \\
\;\;\textsc{\textit{Hybrid Expert}} & $1.50$ & $0.00$ & $0.00$ & $99.50$ & $99.50$ & $98.83$ & $99.00$ \\
\;\;\textsc{\textit{Global Expert}} & $19.67$ & $0.00$ & $0.00$ & $100.00$ & $100.00$ & $97.33$ & $99.50$ \\
\bottomrule
\end{tabular}
\caption{Failure Modes on dresser problems solvable by the hybrid planner}
\label{tab:dresser-hybrid-failure-modes}
\end{table*}

\begin{table*}[h]\centering\footnotesize
\begin{tabular}{@{}lccccccc@{}}
\toprule
& & & & \multicolumn{4}{c}{$\%$ Within} \\
\cmidrule(lr){5-8}
& {$\%$ Env. Coll.} & {$\%$ Self Coll.} & {$\%$ Jnt Viol.} & {1\si{cm}} & {5\si{cm}} & {\ang{15}} & {\ang{30}}  \\
\midrule
G. Fabrics \cite{VanWyk2022GeometricFG} & $19.50$ & $0.33$ & $0.17$ & $40.00$ & $42.67$ & $64.50$ & $68.17$ \\
STORM \cite{Bhardwaj2021FastJS} & $1.17$ & $0.17$ & $0.50$ & $69.50$ & $72.83$ & $91.00$ & $92.33$ \\
\addlinespace
\mpinet (Ours) \\
\;\;\textsc{\textit{Hybrid Expert}} & $1.83$ & $0.00$ & $0.00$ & $99.67$ & $99.67$ & $98.50$ & $98.67$ \\
\;\;\textsc{\textit{Global Expert}} & $14.50$ & $0.00$ & $0.00$ & $100.00$ & $100.00$ & $96.83$ & $99.17$ \\
\bottomrule
\end{tabular}
\caption{Failure Modes on dresser-problems solvable by both the global and hybrid planners}
\label{tab:dresser-both-failure-modes}
\end{table*}

\begin{table*}\centering\footnotesize
\addtolength{\tabcolsep}{-2pt}
\begin{tabular}{@{}lccccc@{}}
\toprule
& & \multicolumn{3}{c}{Success Rate ($\%$)} \\
\cmidrule(lr){3-5}
& Soln. Time (\si{\second}) & Global & Hybrid & Both & Smooth ($\%$) \\
\midrule
\addlinespace
Global Planner \cite{Strub2020AdaptivelyIT} & $16.01 \pm 0.74$ & $100$ & $95.00$ & $100$ & $28.27$ \\
Hybrid Planner & $6.43 \pm 1.18$ & $69.00$ & $96.33$ & $100$ & $100$ \\
\addlinespace
G. Fabrics \cite{VanWyk2022GeometricFG} & $0.14 \pm 0.07$ & $62.50$ & $85.50$ & $85.83$ & $88.61$ \\
STORM \cite{Bhardwaj2021FastJS} & $3.49 \pm 1.65$ & $73.00$ & $88.33$ & $88.67$ & $43.83$ \\
\addlinespace
MPNets~\cite{Qureshi2019MotionPN} \\
\;\;\textit{Hybrid Expert} & $1.36 \pm 7.98$ & $65.67$ & $94.00$ & $94.50$ & $8.23$ \\
\;\;\textit{Random} & $0.05 \pm 0.05$ & $58.17$ & $85.83$ & $89.67$ & $99.94$ \\
\addlinespace
\mpinet (Ours) \\
\;\;\textsc{\textit{Hybrid Expert}} & $0.33 \pm 0.10$ & $81.67$ & $94.67$ & $95.67$ & $96.83$ \\
\;\;\textsc{\textit{Global Expert}} & $0.33 \pm 0.11$ & $78.00$ & $82.33$ & $86.17$ & $80.67$ \\
\bottomrule
\end{tabular}
\caption{Algorithm performance on tabletop problems sets solvable by planner types. All prior methods use state-information and a oracle collision checker while \mpinet only needs a point cloud}
\label{tab:tabletop-success-rates}
\addtolength{\tabcolsep}{2pt}
\end{table*}

\begin{table*}[h]\centering\footnotesize
\begin{tabular}{@{}lccccccc@{}}
\toprule
& & & & \multicolumn{4}{c}{$\%$ Within} \\
\cmidrule(lr){5-8}
& {$\%$ Env. Coll.} & {$\%$ Self Coll.} & {$\%$ Jnt Viol.} & {1\si{cm}} & {5\si{cm}} & {\ang{15}} & {\ang{30}}  \\
\midrule
G. Fabrics \cite{VanWyk2022GeometricFG} & $1.33$ & $0.50$ & $0.17$ & $77.00$ & $87.83$ & $80.67$ & $83.17$ \\
STORM \cite{Bhardwaj2021FastJS} & $0.50$ & $0.17$ & $0.00$ & $92.00$ & $97.67$ & $79.00$ & $83.83$ \\
\addlinespace
\mpinet (Ours) \\
\;\;\textsc{\textit{Hybrid Expert}} & $5.33$ & $1.33$ & $0.00$ & $89.33$ & $96.33$ & $84.33$ & $89.67$ \\
\;\;\textsc{\textit{Global Expert}} & $16.00$ & $0.00$ & $0.00$ & $94.67$ & $98.50$ & $91.33$ & $97.83$ \\
\bottomrule
\end{tabular}
\caption{Failure Modes on tabletop problems solvable by the global planner}
\label{tab:tabletop-global-failure-modes}
\end{table*}

\begin{table*}[h]\centering\footnotesize
\begin{tabular}{@{}lccccccc@{}}
\toprule
& & & & \multicolumn{4}{c}{$\%$ Within} \\
\cmidrule(lr){5-8}
& {$\%$ Env. Coll.} & {$\%$ Self Coll.} & {$\%$ Jnt Viol.} & {1\si{cm}} & {5\si{cm}} & {\ang{15}} & {\ang{30}}  \\
\midrule
G. Fabrics \cite{VanWyk2022GeometricFG} & $1.33$ & $0.00$ & $0.00$ & $97.17$ & $98.00$ & $89.33$ & $89.67$ \\
STORM \cite{Bhardwaj2021FastJS} & $0.17$ & $0.17$ & $0.17$ & $96.67$ & $99.00$ & $90.67$ & $92.17$ \\
\addlinespace
\mpinet (Ours) \\
\;\;\textsc{\textit{Hybrid Expert}} & $0.67$ & $0.00$ & $0.00$ & $97.17$ & $99.50$ & $96.17$ & $98.83$ \\
\;\;\textsc{\textit{Global Expert}} & $11.50$ & $0.00$ & $0.00$ & $97.50$ & $99.50$ & $94.00$ & $98.33$ \\
\bottomrule
\end{tabular}
\caption{Failure Modes on tabletop problems solvable by the hybrid planner}
\label{tab:tabletop-hybrid-failure-modes}
\end{table*}

\begin{table*}[h]\centering\footnotesize
\begin{tabular}{@{}lccccccc@{}}
\toprule
& & & & \multicolumn{4}{c}{$\%$ Within} \\
\cmidrule(lr){5-8}
& {$\%$ Env. Coll.} & {$\%$ Self Coll.} & {$\%$ Jnt Viol.} & {1\si{cm}} & {5\si{cm}} & {\ang{15}} & {\ang{30}}  \\
\midrule
G. Fabrics \cite{VanWyk2022GeometricFG} & $1.33$ & $0.00$ & $0.00$ & $97.33$ & $98.50$ & $89.50$ & $89.83$ \\
STORM \cite{Bhardwaj2021FastJS} & $0.17$ & $0.17$ & $0.17$ & $97.17$ & $99.33$ & $90.50$ & $91.83$ \\
\addlinespace
\mpinet (Ours) \\
\;\;\textsc{\textit{Hybrid Expert}} & $0.50$ & $0.00$ & $0.00$ & $97.33$ & $99.50$ & $96.33$ & $98.33$ \\
\;\;\textsc{\textit{Global Expert}} & $8.67$ & $0.17$ & $0.00$ & $97.00$ & $99.67$ & $95.83$ & $98.17$ \\
\bottomrule
\end{tabular}
\caption{Failure Modes on tabletop problems solvable by both the global and hybrid planners}
\label{tab:tabletop-both-failure-modes}
\end{table*}

\end{document}